\documentclass[sigconf,natbib=true,anonymous=false]{acmart}

\usepackage{booktabs}
\usepackage{multirow}
\usepackage{caption}
\usepackage{subcaption}
\usepackage{CJKutf8}

\newtheorem{definition}{Definition}[]
\newtheorem{problem}{Problem}[]

\AtBeginDocument{%
  }

\copyrightyear{2023}
\acmYear{2023}
\setcopyright{acmlicensed}\acmConference[SIGIR '23]{Proceedings of the 46th
International ACM SIGIR Conference on Research and Development in
Information Retrieval}{July 23--27, 2023}{Taipei, Taiwan}
\acmBooktitle{Proceedings of the 46th International ACM SIGIR Conference on
Research and Development in Information Retrieval (SIGIR '23), July 23--27,
2023, Taipei, Taiwan}
\acmPrice{15.00}
\acmDOI{10.1145/3539618.3591728}
\acmISBN{978-1-4503-9408-6/23/07}

\begin{document}

\title{MGeo: Multi-Modal Geographic Language Model Pre-Training}

\author{Ruixue Ding}
\authornote{Both authors contributed equally to this research.}
\email{ada.drx@alibaba-inc.com}
\author{Boli Chen}
\authornotemark[1]
\email{boli.cbl@alibaba-inc.com}
\affiliation{%
    \institution{Damo Academy,}
    \institution{Alibaba Group,}
    \city{Hangzhou}
    \country{China}
}

\author{Pengjun Xie}
\email{chengchen.xpj@alibaba-inc.com}
\author{Fei Huang}
\email{f.huang@alibaba-inc.com}
\affiliation{%
    \institution{Damo Academy,}
    \institution{Alibaba Group,}
    \city{Hangzhou}
    \country{China}
}

\author{Xin Li}
\email{beilai.bl@alibaba-inc.com}
\author{Qiang Zhang}
\email{muxi.zq@alibaba-inc.com}
\author{Yao Xu}
\email{xuenuo.xy@alibaba-inc.com}
\affiliation{%
    \institution{Gaode Map,}
    \institution{Alibaba Group,}
    \city{Beijing}
    \country{China}
}

\begin{abstract}
    Query and point of interest~(POI) matching is a core task in location-based services~(LBS), \textit{e.g.}, navigation maps. It connects users' intent with real-world geographic information. Lately, pre-trained language models~(PLMs) have made notable advancements in many natural language processing~(NLP) tasks. To overcome the limitation that generic PLMs lack geographic knowledge for query-POI matching, related literature attempts to employ continued pre-training based on domain-specific corpus. However, a query generally describes the \textit{geographic context}~(GC) about its destination and contains mentions of multiple geographic objects like nearby roads and regions of interest~(ROIs). These diverse geographic objects and their correlations are pivotal to retrieving the most relevant POI. Text-based single-modal PLMs can barely make use of the important GC and are therefore limited. In this work, we propose a novel method for query-POI matching, namely Multi-modal Geographic language model~(MGeo), which comprises a geographic encoder and a multi-modal interaction module. Representing GC as a new modality, MGeo is able to fully extract multi-modal correlations to perform accurate query-POI matching. Moreover, there exists no publicly available query-POI matching benchmark. Intending to facilitate further research, we build a new open-source large-scale benchmark for this topic, \textit{i.e.}, Geographic TExtual Similarity~(GeoTES). The POIs come from an open-source geographic information system~(GIS) and the queries are manually generated by annotators to prevent privacy issues. Compared with several strong baselines, the extensive experiment results and detailed ablation analyses demonstrate that our proposed multi-modal geographic pre-training method can significantly improve the query-POI matching capability of PLMs with or without users' locations. Our code and benchmark are publicly available at \url{https://github.com/PhantomGrapes/MGeo}.
\end{abstract}



\maketitle

\begin{figure}[t]
    \centering
    \includegraphics[width=.72\linewidth]{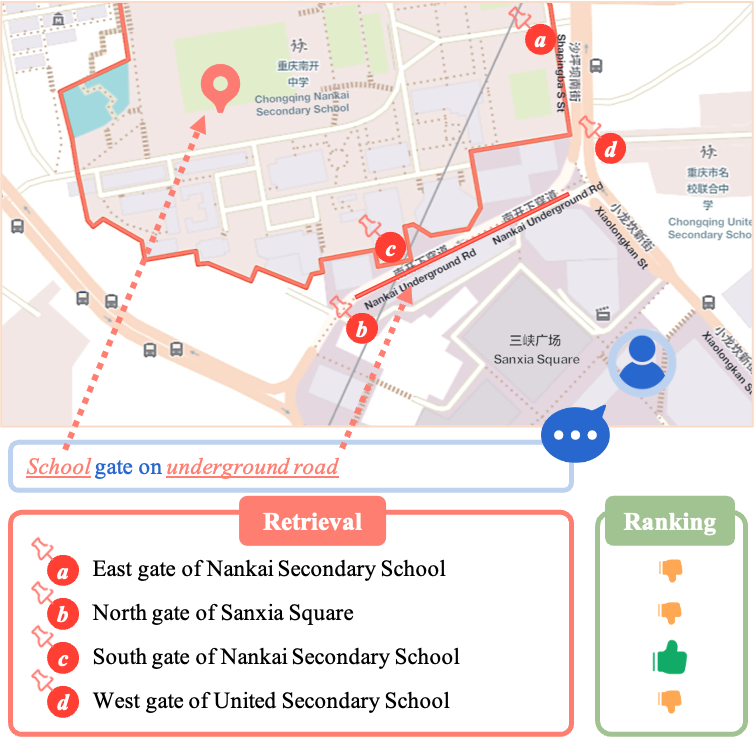}
    \caption{A typical query-POI matching procedure.}
    \label{fig:example}
\end{figure}

\section{Introduction}

As an essential function of location-based services~(LBS) like navigation maps~(\textit{e.g.}, Google Maps), ride-hailing applications~(\textit{e.g.}, Uber), and food delivery platforms~(\textit{e.g.}, Uber Eats), query and point of interest~(POI) matching aims to find a list of candidate POIs based on users' specific or implicit intent. The candidate results are crucial for providing users with real-world geographic information, which directly impacts the navigation, routing, and ordering process. Therefore, effective and accurate query-POI matching is indispensable for delivering a satisfactory user experience. A typical query-POI matching procedure is illustrated in Figure~\ref{fig:example}, which consists of a two-stage retrieve-then-rank pipeline~\cite{10.1145/3404835.3462812, Zhao_Peng_Wu_Chen_Yu_Zheng_Ma_Chai_Ye_Qie_2019}. In specific, given a query, the lightweight retriever first produces an initial set of candidate POIs by searching a massive database, then the ranker sorts the most relevant candidate. This kind of architecture is widely adopted in information retrieval~(IR) systems on account of the efficiency-effectiveness trade-off.

Recent literature on natural language processing~(NLP) as well as IR shows a flourishing advancement of pre-trained language models~(PLMs), notably in semantic textual similarity~(STS) and open-domain question answering~(QA)~\cite{10.1145/3404835.3463034, 10.1145/3404835.3462804, karpukhin-etal-2020-dense}. Continued self-supervised training on domain-specific corpus is shown to be effective for adapting generic PLMs to other domains~\cite{gururangan-etal-2020-dont}. To improve the capability of PLMs for tasks in LBS, various methods have lately been proposed to inject geographic knowledge based on textual data related to geography and user behavioral data~\cite{10.1145/3534678.3539021, liu-etal-2021-geo-bert, 10.1145/3459637.3481924, DBLP:journals/corr/abs-2203-08565}. Although these methods are better at capturing semantic similarity than generic PLMs for query-POI matching, they can barely make use of the more important circumstantial \textit{geographic context}~(GC), \textit{i.e.}, the diverse geographic objects and their correlations from the geographic information system~(GIS)~(detailed in \textsc{Definition}~\ref{def:gc}). Specifically, the geographic objects consist of roads represented as \textit{lines} and regions of interest~(ROIs) represented as \textit{polygons}, the correlations include \textit{near}, \textit{covered}, and their \textit{relative position}.

A query usually mentions multiple geographic objects in the background of the target POI. Fully capturing the information in the GC is necessary for accurate query-POI matching. For example, given the query "school gate on underground road", as shown in Figure~\ref{fig:example}, several relevant POIs are retrieved. The nearest "underground road" to the user is the "Nankai Underground Rd", and the "Nankai Secondary School" has a gate (\textit{c}) on the "Underground Rd". Therefore, the most matched POI should be the gate (\textit{c}). The problem is that the "Nankai Secondary School" is formally located on the "Shapingba S St" with its main gate (\textit{a}). Its side gate (\textit{c}) is not recorded in the GIS as located on the "Underground Rd". It should also be noticed that the user is currently in the "Sanxia Square", which has a gate (\textit{b}) located on the "Underground Rd". The semantic textual similarity alone is not enough to distinguish these two hard negatives (\textit{a}) and (\textit{b}). Moreover, the gate (\textit{d}) of the "United Secondary School" is the closest school gate to the user. Simply considering the relative position of the user and the POI will match the wrong gate (\textit{d}). Only by taking the entire GC into consideration can we find the correct gate (\textit{c}).

To this end, we propose a novel method that draws on GC for query-POI matching, namely Multi-modal Geographic language model~(MGeo). MGeo bridges the modality gap between semantics and GC. MGeo consists of a geographic encoder and a multi-modal interaction module. The geographic encoder makes use of the GC by representing it as a new modality. The multi-modal interaction module then incorporates the geographic features with the semantics. MGeo makes use of the textual, geographic, and cross-modal interactions between queries and POIs. Since the interaction module is compatible with queries that have no GC, it is optional to provide the users' locations, as many applications may require. As a result, rich correlations among textual and geographic modalities can be fully extracted to ensure the quality of query-POI matching.

In addition, there is no public unencrypted benchmark for query-POI matching mostly due to privacy issues. Large publicly available corpus could lead to many breakthroughs in research, \textit{e.g.}, MS MARCO \cite{DBLP:conf/nips/NguyenRSGTMD16}. Intending to facilitate further research on this topic, develop robust techniques, and track progress, we introduce Geographic TExtual Similarity~(GeoTES), which is an open-source large-scale benchmark for query-POI matching with GC~(detailed in Section \ref{sec:geotes}). The POIs come from the open-source GIS OpenStreetMap~(OSM)\footnote{\url{https://www.openstreetmap.org}}. To prevent privacy issues, the queries are manually generated by annotators thus do not require encryption.

Our major contributions are highlighted as follows:
\begin{itemize}
    \item We formalize the important concept GC for the query-POI matching problem and propose a novel method MGeo that uses geographic encoder to represent it as a new modality.
    \item A multi-modal interaction module is proposed to incorporate the correlations among textual and geographic modalities. It is compatible with queries that have no GC as well.
    \item A new open-source large-scale benchmark GeoTES is built to facilitate further research. The POIs come from an open-source GIS and the queries are manually generated by annotators to prevent privacy issues.
    \item Compared with strong baselines, the experiment results demonstrate that our proposed methods can significantly improve the query-POI matching capability of PLMs, even when no GC is provided for the queries.
\end{itemize}

\section{Related Work}

\subsection{Relevance Model}

Traditional approaches for retrieving documents from large corpus generally use exact term-level matching, \textit{e.g.}, Okapi Best Matching~(BM25)~\cite{DBLP:conf/trec/RobertsonWJHG94}. Despite such heuristic retrievers having low latency via inverted list data structure, their measurement of similarity is only based on document statistics.
Latterly, Deep neural network~(DNN) models have been introduced to IR. For example, Deep Structured Semantic Model~(DSSM)~\cite{10.1145/2505515.2505665} measures the relevance of queries and documents in a semantic vector space by computing their cosine similarity. Along with the success of PLMs in NLP, studies on IR have also made remarkable progress~\cite{10.1145/3404835.3463034, karpukhin-etal-2020-dense, 10.1007/978-3-030-72240-1_26}. On account of the efficiency-effectiveness trade-off, there are two major architectures, \textit{i.e.}, bi-encoder and cross-encoder~\cite{thakur-etal-2021-augmented}. Bi-encoder allows efficient indexing~\cite{reimers-gurevych-2019-sentence, 10.1145/3331184.3331303} and is usually used in the retrieval system. In contrast, cross-encoder concatenates the query and document to perform cross-interaction over all input terms. Although cross-encoder can provide a more accurate estimation of relevance, it needs more computing resources and is usually used only in the ranking system. MGeo can be applied on both bi-encoder and cross-encoder architecture.

\subsection{Multi-Modal Representation Learning}

Following the tremendous success of various pre-training techniques in NLP, a lot of Transformer-based models are proposed for other modalities, such as compute vision~(CV)~\cite{10.1007/978-3-030-58452-8_13, dosovitskiy2021an, DBLP:conf/icml/KimSK21}. Except for single-modal, recent studies also show the derivative models have great potential in multi-modal representation learning~\cite{Sun_2019_ICCV, DBLP:conf/eccv/ChenLYK0G0020, DBLP:conf/icml/KimSK21, 10.1145/3447548.3467206}. For example, CLIP~\cite{pmlr-v139-radford21a} converts classification to a retrieval task and enables zero-shot learning via large-scale multi-modal pre-training. In addition to image, layout of document and table can also be represented as different modalities~\cite{xu-etal-2021-layoutlmv2, 10.1145/3485447.3511972}. In this paper, our proposed MGeo use the geographic encoder to represents GC as a new modality for query-POI matching.

\subsection{Query-POI Matching}

Previous work on query-POI matching generally focuses on modeling the relative position between queries and POIs. Based on DSSM, PALM~\cite{Zhao_Peng_Wu_Chen_Yu_Zheng_Ma_Chai_Ye_Qie_2019} obtains the positional relationship of queries and POIs from coordinate-based and kernel-based location embeddings, and incorporates the relationship with semantic similarity for POI retrieval. STDGAT~\cite{10.1145/3397271.3401159} further takes multiple spatiotemporal factors into consideration via dual graph attention network when quantifying the query-POI relevance. On account of the ubiquity of PLMs in NLP, domain-adaptive pre-training methods have been proposed to inject extralinguistic knowledge into the generic PLMs~\cite{DBLP:journals/corr/abs-2203-08565, liu-etal-2021-geo-bert}. Typically, GeoL~\cite{10.1145/3534678.3539021} makes use of the static geographic knowledge based on user behavior~(search logs), \textit{e.g.}, geocoding~\cite{Goldberg2007FromTT}. Although the domain-adapted PLMs may be better at capturing the semantic similarity than generic PLMs for query-POI matching, they are still limited by ignoring GC in the background.

Moreover, to facilitate further research and promote the development of robust techniques, we also establish a reliable public large-scale query-POI matching benchmark named GeoTES.
\section{Preliminary}
\label{prel}

We first introduce the formal description of the query-POI matching problem, as well as some important definitions related to GC. Table~\ref{table:notation} gives the frequently used notations.

Let $P$ be the set of POIs $p$. $P$ can either contain dozens of candidate POIs or a large number of POIs in the massive database. Each POI $p$ consists of a textual description $t^p$ and its geolocation $l^p$. The textual description of the POI $t^p$ contains its formal address and name. Let $q$ denote a query made by the user. The textual description of the query $t^q$ belongs to three types, \textit{i.e.}, common \textit{address} description, formal \textit{street number} description, and casual \textit{colloquial} description.
The \textit{street number} query contains standard numerical designation for a target POI, while the \textit{address} query does not. The \textit{colloquial} query uses spoken language and may contain colloquial noisy words.
The query's geolocation $l^q$ can be the users' geolocation. When the user searches for another area using the map, $l^q$ is the center location displayed on screen. Furthermore, $l^q$ may or may not be provided. We denote geolocation of a POI or query as $l^{pq}$.

\begin{problem}
\label{prob:prob1}
\textbf{Query-POI matching problem}. Given the POI set $P$ and a user's query $q$ in LBS, we aim to estimate the POI $p \in P$ that best matches the user's intent.
\end{problem}

We define two tasks based on the size of $P$, \textit{i.e.}, \textit{ranking} and \textit{retrieval}. Specifically, for the ranking task, $P$ is a list of candidate POIs with a limited number, where the best-matched one is included. As for the retrieval task, $P$ is the massive database that contains all POIs, and the total number of POIs is large. Since cross-encoder is inefficient for large size of $P$, it only runs on the ranking task. Bi-encoder can run on both the ranking and retrieval tasks.

\begin{definition}
    \textbf{Geographic object}.
    GIS is constructed on spatial data that defines the real-world geometric space. Let $\mathcal{G}$ be the spatial database. Each geographic object $o \in \mathcal{G}$ with $m$ vertices is described as a sequence of geolocation $\{l^o_1, l^o_2, \dots, l^o_m\}$. A geographic object is intrinsically characterized by its ID, absolute position in the map, and shape $o^s \in \{LINE, POLYGON\}$. Specifically, $LINE$ represents the real-world road and $POLYGON$ represents the ROI.
\end{definition}

Here we use $m$ to denote the number of vertices in $o$. Note that given the geolocation of the POI or the query, we can form a list of nearby geographic objects $\{o_1, o_2, \dots, o_n\}$ sorted by distance, \textit{i.e.}, $o_1$ is the nearest geographic object to the POI or query. $n$ is used to denote the number of geographic objects for a geolocation $l^{pq}$. We export OSM to PostGIS\footnote{\url{https://postgis.net/}} and get the Geographic Context~(GC) of a geolocation from it.

\begin{definition}
    \label{def:gc}
    \textbf{Geographic context (GC)}. Given the geolocation $l^{pq}$ of a POI or query, where $l^{pq}$ is represented by a geographic coordinate $(lng, lat)$, the GC is characterized by the correlations between $l^{pq}$ and its $n$ geographic objects $\{o_1, o_2, \dots, o_n\}$. Formally, the relation type $r^t \in \{NEAR, COVERED\}$ indicates whether $l^{pq}$ is inside $o_i$ or at a distance. The relative position $r^p$ depicts a more detailed positional relationship between $l^{pq}$ and $o_i$.
\end{definition}

When searching for a target POI, a user usually explores the nearby circumstantial spatial data and mentions multiple related geographic objects in the query. The intrinsic characteristics of geographic objects are also important to extract GC information. Therefore, modeling the intrinsic characteristics of geographic objects is pivotal to capturing correlations in GC and ensuring the quality of query-POI matching. The methods used to encode GC are detailed in Section~\ref{sec:encoding}.

\begin{table}[t]
    \footnotesize
    \caption{Table of notations.}
    \label{table:notation}
    \begin{tabular}{ll}
        \toprule
        Notation       & Description                                              \\ \midrule
        $P, p$         & The POI set and a POI.                                   \\
        $q$            & A query given by the user.                               \\
        $o$            & A geographic object.                                     \\
        $o^s$          & The shape of geographic object, $\in \{LINE, POLYGON\}$. \\
        $o^m$          & The position of $o$ in the map.                          \\
        $r^t$          & The relation type $\in \{NEAR, COVERED\}$.               \\
        $r^p$          & The relative position.                                   \\
        $t$            & The textual description of POI or query.                 \\
        $l=(lng, lat)$ & The geolocation represented by longitude and latitude.   \\
        $l^{pq}$       & The geolocation of a POI or query.                       \\
        $l^{o}$        & A vertex of $o$.                                         \\
        $\tilde{o}$    & The rectangle that approximates the shape of $o$.        \\ \bottomrule
    \end{tabular}
\end{table}

\section{The GeoTES Benchmark}
\label{sec:geotes}

\begin{CJK*}{UTF8}{gbsn}
    \begin{table*}[t]
        \caption{Examples from GeoTES. We only show the positive POI and one geographic object here for simplicity. Every query can have multiple negative POIs similar to the positive POI. Every query and candidate POI can have GC of multiple related geographic objects. Note that query GC is optional to simulate the absence of user location.}
        \label{table:example}
        \centering
        \small
        \begin{tabular}{llll}
            \toprule
            \cmidrule(r){1-3}
                                                               &                                          & Query                                     & Positive POI                          \\ \midrule
            \multirow{2}{*}{Text}                              &                                          & 滨海大厦对面江汉路                        & 6号线江汉路(地铁站)                   \\
                                                               &                                          & Jianghan Road opposite to Binhai Building & Line 6 Jianghan Road (subway station) \\ \midrule
            \multicolumn{2}{l}{Location~(Longitude, Latitude)} & (120.20435566081441, 30.210982121527547) & (120.20039749738959, 30.20525878443309)                                           \\ \midrule
            \multirow{3}{*}{Geographic Object}                 & ID                                       & 41935                                     & 42599                                 \\
                                                               & Shape                                    & $POLYGON$                                 & $LINE$                                \\
                                                               & Map Position                             & {[}1324, 1341, 1325, 1342{]}              & {[}1322, 1335, 1323, 1336{]}          \\ \midrule
            \multirow{2}{*}{Geographic Context}                & Relation                                 & $NEAR$                                    & $COVERED$                             \\
                                                               & Relative Position                        & {[}31, 24, 22, 22{]}                      & {[}27, 27, 18, 19{]}                  \\ \bottomrule
        \end{tabular}
    \end{table*}
\end{CJK*}

\begin{table}[t]
    \caption{Statistics of different query types.}
    \label{table:query_types}
    \centering
    \small
    \begin{tabular}{lc}
        \toprule
        Query Type & \# Query \\ \midrule
        Address    & 81,286   \\
        Street No. & 6,013    \\
        Colloquial & 2,701    \\ \midrule
        Total      & 90,000   \\ \bottomrule
    \end{tabular}
\end{table}

\begin{table}[t]
    \caption{Statistics of geographic objects, which describe the average number of shapes with corresponding relation to queries and POIs.}
    \label{table:gis_stats}
    \centering
    \small
    \begin{tabular}{lcccc}
        \toprule
              & \multicolumn{2}{c}{$LINE$} & \multicolumn{2}{c}{$POLYGON$}                      \\ \midrule
              & $NEAR$                     & $COVERED$                     & $NEAR$ & $COVERED$ \\ \midrule
we        Query & 4.4                        & 0.005                         & 14.2   & 0.7       \\
        POI   & 3.7                        & 0.003                         & 10.4   & 0.6       \\ \bottomrule
    \end{tabular}
\end{table}

In this section, we introduce our proposed large-scale benchmark GeoTES, which stands for Geographic TExtual Similarity. It is the first open-source benchmark for query-POI matching. The POIs are obtained from the open-source OSM and the queries are manually generated by annotators to prevent privacy issues.

In this version of GeoTES, all the POIs are located in Hangzhou and use Chinese text. Table~\ref{table:example} give examples from the benchmark. Since geographic objects in GIS and GC are language-agnostic, MGeo can be easily applied to multilingual situations. Section \ref{sec:encoding} details how to obtain map position and relative position. A query is equipped with a positive POI, and negative POIs with a limited number are provided for the ranking task. \footnote{Benchmark is available at \url{https://modelscope.cn/datasets/damo/GeoGLUE}.}

\subsection{Annotation Process}

We recruited 20 annotators and 4 experienced experts to annotate three types of queries based on POIs defined in Section~\ref{prel}. Table~\ref{table:query_types} gives the statistics of these query types, which follows the distribution of our online LBS. In OSM, each POI comes with a geographic location under the WGS84 coordinate system.\footnote{\url{https://wiki.openstreetmap.org/wiki/Converting_to_WGS84}} Neighbouring POIs of the OSM POIs from several open-accessed map services are selected by the annotators to enrich the diversity of POI description and also serve as hard negatives. To simulate the queries' location in real scenes, the annotators are asked to randomly select a location within 1km of corresponding POI for 50\% of the queries and randomly select a location in the city for the rest queries. All the annotators have adequate linguistic knowledge and educational/cultural background to produce appropriate queries. To eliminate biases during the annotation process, they are instructed with detailed annotation principles. A quality inspector verifies the annotations and confirms that each query has exactly one positive POI.

\begin{table}[t]
    \caption{Statistics of train/dev/test splits.}
    \label{table:split_stats}
    \centering
    \small
    \begin{tabular}{lcc}
        \toprule

                         & \# Query                & \# Candidate POI \\\midrule
        Train            & 50,000                  & 20               \\\midrule
        Dev              & 20,000                  & 40               \\\midrule
        Test (Ranking)   & \multirow{2}{*}{20,000} & 40               \\
        Test (Retrieval) &                         & 2,849,754        \\ \bottomrule
    \end{tabular}
\end{table}

\subsection{Benchmark Statistics}

GeoTES has a total number of 90,000 queries with an average length of 17.2 and 2,849,754 POIs with an average length of 13.7. We extract the geographic surrounding objects for the queries and POIs from OSM. There are 21,950 lines and 65,722 polygons in our extracted geographic objects. Table~\ref{table:gis_stats} gives the averaged number of different shapes and relations, each query and POI has more $NEAR$ relation and more relations to polygons~(ROIs) than lines~(roads). As shown in Table~\ref{table:split_stats}, the benchmark is randomly split in to train, development, and test sets. For the train, development, and ranking test sets, we provide a list of candidate POIs and ensure that one exact matched positive POI is contained. The retrieval test set use the same queries as the ranking test set while no candidate POI list should be provided. GeoTES is thus a reliable and challenging benchmark for evaluating both retrieval and ranking models.

\section{Methods}

\begin{figure*}[t]
    \centering
    \includegraphics[width=.8\linewidth]{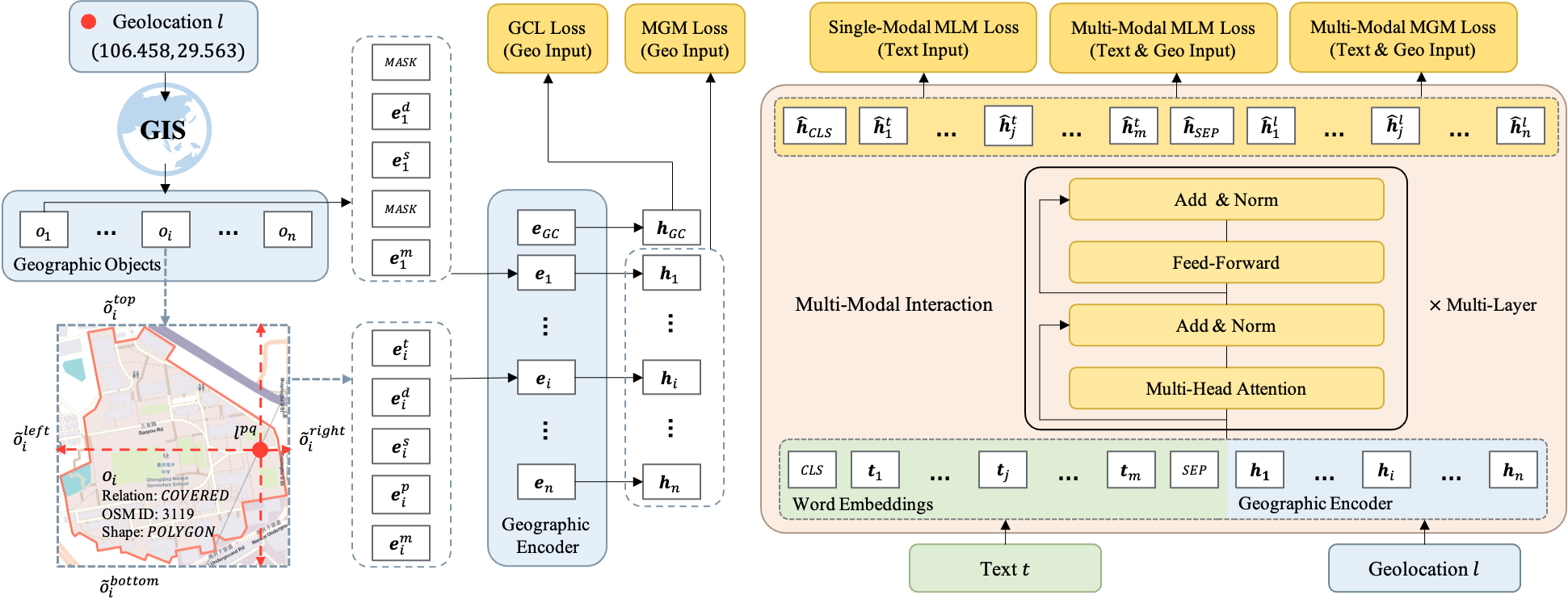}
    \caption{Architecture of MGeo. Left part shows encoding and pre-training process of geographic encoder and right part shows the multi-modal pre-training process of MGeo. Word embeddings of text $\mathbf{t}$ and GC representations of geographic encoder $\mathbf{h}$ are concatenated together and fed to multi-modal interaction module, which produces final representations $\hat{\mathbf{h}}^t$ for each text token and $\hat{\mathbf{h}}^l$ for each geographic object.}
    \label{fig:mgeo}
\end{figure*}

In this section, we present the detailed architecture and pre-training process of MGeo. Following state-of-the-art multi-modal methods~\cite{DBLP:conf/icml/KimSK21, DBLP:journals/corr/abs-2103-00823, DBLP:conf/eccv/ChenLYK0G0020}, MGeo is composed of a geographic encoder and a multi-modal interaction module, as shown in Figure~\ref{fig:mgeo}. The full training process of MGeo consists of three steps: (1)~we first train geographic encoder alone to learn representations of GC, and the trained geographic encoder is fixed in the following stages; (2)~text-GC pairs are then used to pre-train MGeo in a multi-modal way, by modeling geographic objects along with text and pre-training with massive text-GC pairs, and MGeo successfully aligns these two modals into a same latent space; (3) MGeo is lastly fine-tuned on ranking and retrieval tasks and gains significant improvements.

\subsection{Geographic Encoder}
\label{sec:geo_enc}

The geolocation alone is meaningless unless it has GC. Taking a geolocation $l$ as input, geographic encoder maps the GC as a new modality to dense representations, which contains features of the surrounding geographic objects $\{o_1, o_2, \dots, o_n\}$.

\subsubsection{Encoding}
\label{sec:encoding}

Geographic encoder can extract the correlations between query/POI geolocation~(point) and their surrounding geographic objects~(line or polygon). Geographic encoder respectively represents the intrinsic characteristics of geographic objects~(\textit{i.e.}, ID, shape, and map position), the relations~(\textit{i.e.}, $NEAR$ or $COVERED$), and the relative position as embeddings.

\paragraph{ID}
To extract the intrinsic features of geographic objects, the OSM IDs are mapped to embeddings in a similar way to word embeddings. %
The ID embeddings of $o_i$ are denoted as $\mathbf{e}^d_i$.

\paragraph{Shape}
A one-hot function is used to encode the categorical shape type $o^s_i$ as a numeric array and to obtain its corresponding embeddings $\mathbf{e}^s_i$. The shape type embeddings are denoted as $\mathbf{e}^s_i$.

\paragraph{Map position}
The absolute position of $o_i$ in the map $\mathbf{e}^m_i$ is pivotal to distinguishing itself from other geographic objects. The \textbf{entire} map area as a rectangle is split into a $N \times N$ grid to obtain its scale factors $s_{lng}$ and $s_{lat}$ for longitude and latitude respectively:
\begin{eqnarray}
    s_{lng}= \frac{lng^{m_{right}} - lng^{m_{left}}}{N}, s_{lat}= \frac{lat^{m_{top}} - lat^{m_{bottom}}}{N},
\end{eqnarray}
where $lng^{m_{right}}$ denotes longitude of the map's right side and so on. The position of $o_i$ in the map $o^m_i$ can thus be calculate with the scale factors. For example, $o^{m_{left}}_i$ and $o^{m_{bottom}}_i$ are calculated as:
\begin{eqnarray}
    o^{m_{left}}_i \quad =& \lfloor \frac{\tilde{o}^{left}_i - lng^{m_{left}}}{s_{lng}} \rfloor &\in \mathbb{N}, \\
    o^{m_{bottom}}_i \quad =& \lfloor \frac{\tilde{o}^{bottom}_i - lat^{m_{bottom}}}{s_{lat}} \rfloor &\in \mathbb{N}.
\end{eqnarray}
The discretized position feature of $o_i$ in the map is then encoded as $\mathbf{e}^m_i = \{\mathbf{e}^{m_{left}}_i, \mathbf{e}^{m_{bottom}}_i, \mathbf{e}^{m_{right}}_i, \mathbf{e}^{m_{top}}_i\}$.

\paragraph{Relation}
Similar to the shape type, a one-hot function is used to encode the relation type $r^t_i$ of $o_i$ as a numeric array and to obtain its corresponding embeddings $\mathbf{e}^t_i$.

\paragraph{Relative Position}
To simplify the relative position $r^p_i$, we form a rectangle $\tilde{o_i}$ of similar size to approximate the shape of $o_i$. Each side of the substituted rectangle~(left, bottom, right, and top) is defined as:
\begin{eqnarray}
    \tilde{o}^{left}_i &=& min\bigl(\{lng^{o_i}_j\}_{j \in \{1, \dots, m_i\}}\bigr), \\
    \tilde{o}^{bottom}_i &=& min\bigl(\{lat^{o_i}_j\}_{j \in \{1, \dots, m_i\}}\bigr), \\
    \tilde{o}^{right}_i &=& max\bigl(\{lng^{o_i}_j\}_{j \in \{1, \dots, m_i\}}\bigr), \\
    \tilde{o}^{top}_i &=& max\bigl(\{lat^{o_i}_j\}_{j \in \{1, \dots, m_i\}}\bigr),
\end{eqnarray}
where $lng$ denotes longitude and $lat$ denotes latitude of $l^{o_i}_j$ for simplicity. The relative position $r^p_i = \{r^{p_{left}}_i, r^{p_{bottom}}_i, r^{p_{right}}_i, r^{p_{top}}_i\}$ is then calculated by the normalized distances between $l^{pq}$ and each side of the $\tilde{o_i}$. For example, $r^{p_{left}}_i$ is calculated as:
\begin{eqnarray}
    r^{p_{left}}_i = sgn(lng^{pq} - \tilde{o}^{left}_i) * min \bigl(k, \lfloor k \frac{ |lng^{pq} - \tilde{o}^{left}_i| }{\tilde{o}^{right}_i - \tilde{o}^{left}_i} \rfloor \bigr) + k,
\end{eqnarray}
where $sgn(\cdot)$ is the sign function, and $\lfloor\cdot\rfloor$ is the floor function that outputs the greatest integer less than or equal to a number. $k \in \mathbb{N}$ is a discretization factor that maps the relative distance ratio to a discrete number. As a result, we have $r^{p_{left}}_i \in \{0, \dots, 2k\}$. The discretized relative position feature is then encoded as $\mathbf{e}^p_i = \{\mathbf{e}^{p_{left}}_i, \mathbf{e}^{p_{bottom}}_i, \mathbf{e}^{p_{right}}_i, \mathbf{e}^{p_{top}}_i\}$.

Finally, the geographic encoder sums these features of $o_i$ up as:
\begin{eqnarray}
    \mathbf{e}_i = \mathbf{e}_i^d + \mathbf{e}_i^s + \sum \mathbf{e}^m_i + \mathbf{e}_i^t + \sum \mathbf{e}_i^p
\end{eqnarray}
The intrinsic characteristics of geographic objects are described by the three components~($\mathbf{e}^d$, $\mathbf{e}^s$, and $\mathbf{e}^m$). $\mathbf{e}^d$ is the unique identifier of a geographic object, $\mathbf{e}^s$ distinguishes road from ROI, $\mathbf{e}^m$ depicts the positional relation among different geographic objects. The other two components~($\mathbf{e}^t$ and $\mathbf{e}^p$) describe correlations between geolocation and geographic objects. After encoding surrounding geographic objects as a sequence $\{\mathbf{e}_1, \dots, \mathbf{e}_m\}$, geographic encoder employs multi-layer bidirectional transformers~\cite{DBLP:conf/nips/VaswaniSPUJGKP17} to learn interactions among them. Following previous work~\cite{thakur-etal-2021-augmented}, a $GC$ token is prepended at the beginning like the $CLS$ token. The outputs of geographic encoder are therefore denoted as $\{\mathbf{h}_{GC}, \mathbf{h}_1, \dots, \mathbf{h}_m\}$.

\begin{figure}[t]
    \centering
    \begin{subfigure}[t]{.49\linewidth}
        \centering
        \includegraphics[width=1.\linewidth]{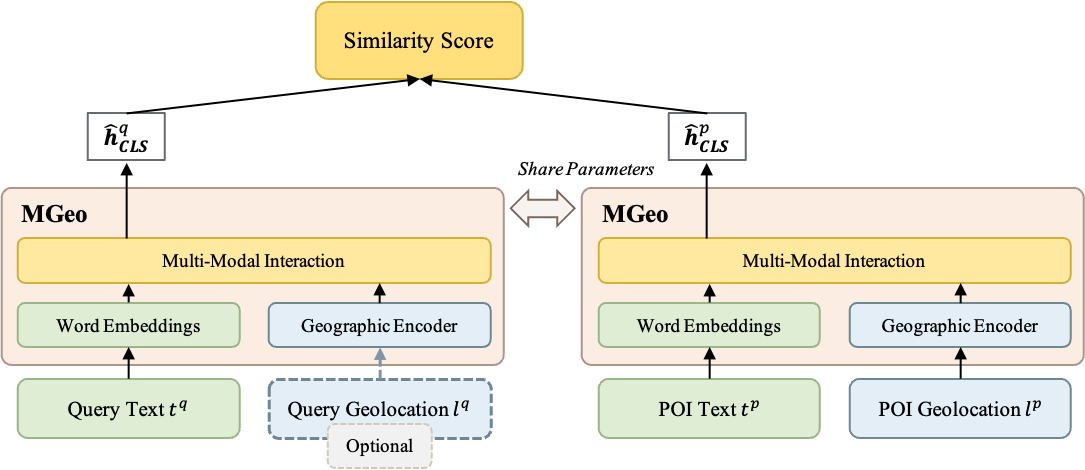}
        \caption{Bi-Encoder.}
        \label{fig:bi_encoder}
    \end{subfigure}
    \begin{subfigure}[t]{.49\linewidth}
        \centering
        \includegraphics[width=1.\linewidth]{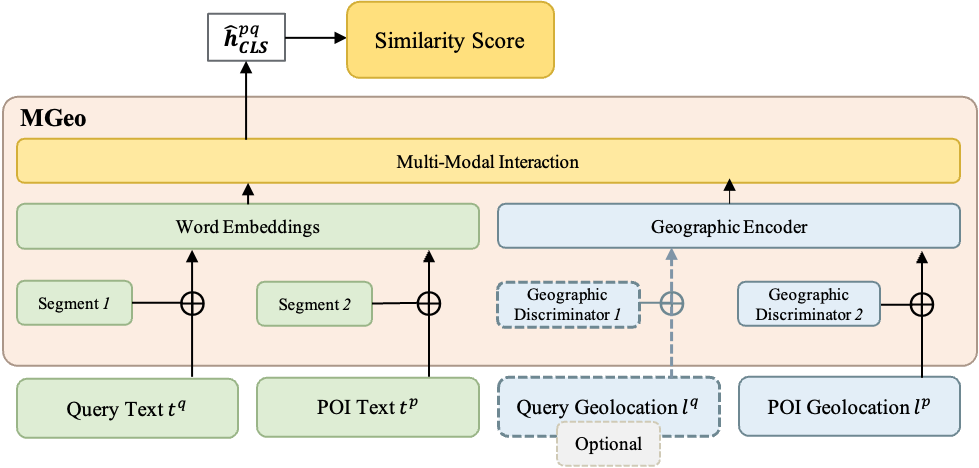}
        \caption{Cross-Encoder.}
        \label{fig:cross_encoder}
    \end{subfigure}
    \caption{MGeo can use both (A) bi-encoder and (B) cross-encoder architectures to measure relevance between query and POI. Dashed line indicates that geolocation of query is optional. $\oplus$ denotes element-wise addition.}
    \label{fig:bi_cross_encoder}
\end{figure}

\subsubsection{Training}
We design two tasks to train geographic encoder and it is fixed in later uses, \textit{i.e.}, masked geographic modeling~(MGM) and geographic contrastive learning~(GCL).

\paragraph{MGM}
Like the widely use masked language modeling~(MLM)~\cite{DBLP:conf/naacl/DevlinCLT19}, MGM aims at predicting masked geographic features, \textit{i.e.}, OSM IDs, geometric types, each side of the substituted rectangle, relation types, and relative positions. The MGM loss $L_{MGM}$ is calculated by summing up the masked loss of all features.

\paragraph{GCL}
This task is related to multiple geolocations $\{l^{pq}_1, \dots, l^{pq}_{bs}\}$ in a batch of size $bs$. We begin with the definition of the real-world geographic distance matrix $\mathbf{H} \in \mathbb{R}^{bs \times bs}$ defined as:
\begin{eqnarray}
    \mathbf{H}_{i,j} = \sigma \bigl( -\| haversine(l^{pq}_i,l^{pq}_j) \|_\mathcal{N} \bigr), i, j \in \{1, \dots, bs\}, i \neq j,
\end{eqnarray}
where $haversine$ is the haversine function~\cite{22de05fbcd8b496bb6da77bced3e3ba6} that calculates spherical distance between geolocations, $\|\cdot\|_\mathcal{N}$ is gaussian normalization function, and $\sigma$ is sigmoid function that maps distance to range $(0, 1)$. As the latent distance between embeddings in the output space should correspond to their real-world geographic distance, we use $h_{GC}$ as the representation of geolocation $l^{pq}$ with GC and calculate the latent distance matrix $\tilde{\mathbf{H}} \in \mathbb{R}^{bs \times bs}$ as:
\begin{eqnarray}
    \tilde{\mathbf{H}}_{i,j} = \langle \|\mathbf{h}_{GC}^i\|_{L^2}, \|\mathbf{h}_{GC}^j\|_{L^2} \rangle
\end{eqnarray}
where $\langle\cdot\rangle$ denotes the doc-product function and $\|\cdot\|_{L^2}$ is $L^2$ normalization function. We use KL-divergence to measure the similarity between $\mathbf{H}$ and $\tilde{\mathbf{H}}$. GCL loss $L_{GCL}$ is then calculated by:
\begin{eqnarray}
    L_{GCL} = \sum_{i = 1}^{bs} D_{KL} \bigl( softmax(\mathbf{H}_i)\ \|\ softmax(\tilde{\mathbf{H}}_i) \bigr)
\end{eqnarray}
where $D_{KL} (\cdot\ \|\ \cdot)$ denotes the KL-divergence, and the $softmax$ function is applied to transform $\mathbf{H}_i$ and $\tilde{\mathbf{H}}_i$ to a distribution.

The training loss $L_g$ of geographic encoder is thus calculated by:
\begin{eqnarray}
    L_g = L_{MGM} + L_{GCL}
\end{eqnarray}
Using such an training process, geographic encoder is capable of modeling GC in a given GIS.

\begin{table*}[t]
    \caption{Ranking results of bi-encoder and cross-encoder PLMs. Bold indicates the best of each column.}
    \label{table:ranking}
    \centering
    \small
    \begin{tabular}{llcccccccc}
        \toprule
                   &                                             & \multicolumn{4}{c}{Bi-Encoder} & \multicolumn{4}{c}{Cross-Encoder}                                                                                                       \\ \midrule
        PLM        &                                             & Recall@1                       & Recall@3                          & Recall@5       & MRR@5          & Recall@1       & Recall@3       & Recall@5       & MRR@5          \\ \midrule
        BERT       &                                             & 58.83                          & 79.40                             & 86.24          & 69.60          & 81.52          & 91.11          & 94.10          & 86.53          \\
        RoBERTa    &                                             & 68.52                          & 85.41                             & 90.25          & 76.15          & 83.20          & 93.09          & 95.77          & 88.30          \\
        ERNIE      &                                             & 58.18                          & 81.86                             & 88.96          & 70.43          & 81.82          & 91.79          & 94.73          & 87.01          \\
        StructBERT &                                             & 69.09                          & 86.29                             & 91.09          & 77.96          & 83.51          & 93.21          & 95.67          & 88.53          \\ \midrule
        BERT       & \multirow{2}{*}{DA}                         & 72.49                          & 89.18                             & 93.48          & 81.03          & 83.24          & 92.92          & 95.63          & 88.25          \\
        StructBERT &                                             & 74.30                          & 89.78                             & 94.06          & 82.28          & 83.65          & 93.33          & 95.92          & 88.61          \\ \midrule
        BERT       & \multirow{2}{*}{\begin{tabular}[c]{@{}l@{}}MGeo  \\ w/o query GC\end{tabular}} & 74.86                          & 90.61                             & 94.53          & 82.93          & 85.11          & 94.42          & 96.75          & 89.86          \\
        StructBERT &                                             & 75.37                          & 89.99                             & 93.96          & 82.89          & 84.72          & 93.85          & 96.16          & 89.40          \\ \midrule
        BERT       & \multirow{2}{*}{MGeo}                       & 76.04                          & \textbf{91.24}                    & \textbf{95.18} & \textbf{83.85} & 85.89          & 95.48          & 97.48          & 90.74          \\
        StructBERT &                                             & \textbf{76.07}                 & 90.68                             & 94.50          & 83.57          & \textbf{86.49} & \textbf{95.55} & \textbf{97.62} & \textbf{91.10} \\ \bottomrule
    \end{tabular}
\end{table*}

\begin{table}[t]
    \caption{Model sizes of pre-trained and fine-tuned models.}
    \label{table:model_size}
    \centering
    \small
    \begin{tabular}{lcc}
        \toprule
                  & Pre-Train & Fine-Tune \\ \midrule
        BERT-DA   & 118M      & 102M      \\
        BERT-MGeo & 213M      & 129M      \\ \bottomrule
    \end{tabular}
\end{table}

\subsection{Multi-Modal Pre-Training}

The input of MGeo pre-training is a pair of text and geolocation ($t$, $l$). The pre-training data can come from diverse sources, \textit{e.g.}, click of users or position of delivery clerks. The multi-modal training aims at aligning these two modals into one latent space. Word embeddings are used to map text into a sequence of vectors. The geographic encoder provides the GC embeddings given $l$. The two embeddings are then concatenated together and fed into multi-layer bidirectional Transformers.

We use three tasks to learn interaction between GC and text, \textit{i.e.}, single-modal MLM, multi-modal MLM, and multi-modal MGM. These tasks are trained in turns. Single-modal MLM is the original MLM task used in BERT, which randomly masks and replaces the input text with $MASK$ token. The outputs of geographic encoder are removed for single-modal MLM. While multi-modal MLM predicts the masked token relying on the entire GC and part of textual information. Multi-modal MGM randomly masks and replaces the input geographic features with $MASK$ and predicts them relying on entire textual information and part of GC.

\subsection{Relevance Measurement}

MGeo can use both bi-encoder and cross-encoder architectures, as shown in Figure~\ref{fig:bi_cross_encoder}. Bi-encoder encodes query and POI separately for efficiency issues. It can be used in both retrieval and ranking phases. In practice, the GC of a POI or query is encoded by geographic encoder. Since user location is not always available due to privacy issues or limited hardware, the GC of query can be absent. The outputs are then concatenated with word embeddings. Transformer-based multi-modal interaction module then produces hidden states as final representations. We compute the similarity score of a query and POI pair by the cosine similarity between their $CLS$ representations, \textit{i.e.}, $\hat{\mathbf{h}}^p$ and  $\hat{\mathbf{h}}^p$. Bi-encoder calculates similarity scores between a query and all the POIs for retrieval task. %

Different from bi-encoder, cross-encoder concatenates every query-POI pair together before being fed to multi-modal interaction module. Cross-encoder allows fine-grained token-level interaction between query and POI, it usually provides a more accurate estimation of relevance but is less efficient. Therefore, cross-encoder is only used in ranking phase as usual. The GC of query or POI is encoded separately by geographic encoder. The GC of query is also optional. We concatenate query textual embeddings, POI textual embeddings, query GC embeddings (optional), and POI GC embeddings together, which are then fed to multi-modal interaction module. Particularly, we use geographic discriminator to facilitate geographic comparison between GC of query and POI. Geographic discriminator adds embeddings to outputs of geographic encoder to distinguish query GC from POI GC. Like the segment embeddings in BERT, embeddings of geographic discriminator are randomly initialized and trainable. We fed the hidden states of $CLS$ $\hat{\mathbf{h}}^{pq}_{CLS}$ to a multi-layer perceptron~(MLP) to produce similarity scores.

\section{Experiments}

In this section, we compare the proposed MGeo with several strong baselines on GeoTES.

\subsection{Setup}

\subsubsection{Tasks}
The experiments are conducted on two tasks, \textit{i.e.}, \textit{ranking} and \textit{retrieval}. The two tasks use the same train, development, and test sets as shown in Table~\ref{table:split_stats}. A list of candidate POIs that contains the positive POI is provided for the ranking test set. Both bi-encoder and cross-encoder are evaluated on ranking task. Since retrieval task requires searching the full POI corpus, and cross-encoder needs too much computing resources to complete retrieval task, only bi-encoder is evaluated on retrieval task.

\subsubsection{Evaluation metrics}
Following previous IR work~\cite{qu-etal-2021-rocketqa}, we use Recall and Mean Reciprocal Rank~(MRR) at top $k$ ranks to evaluate the performance on both tasks. Recall@$k$ calculates the proportion of queries that have the positive POI contained in the top-$k$ candidates, and MRR@$k$ calculates the averaged reciprocal of the rank at which the positive POI is placed. We report the evaluation scores on the test set of models that perform best on the development set during training.

\subsubsection{PLM Baselines} We first evaluate the performance of four widely used PLMs with the base model size on GeoTES, including BERT~\cite{DBLP:conf/naacl/DevlinCLT19}, RoBERTa\footnote{\url{https://huggingface.co/clue/roberta_chinese_base}}~\cite{DBLP:journals/corr/abs-1907-11692}, ERNIE 3.0~\cite{DBLP:journals/corr/abs-2107-02137}, and StructBERT~\cite{Wang2020StructBERT:}. We further apply domain-adaptive pre-training techniques (DA) on BERT and another top-performing model. DA is a widely used single-modal pre-training baseline \cite{gururangan-etal-2020-dont}. For
a fair comparison, domain corpus used in DA is the same as that used in our proposed multi-modal geographic pre-training (MGeo), except that MGeo has additional GC along with query and POI.

\subsubsection{Hyperparameters}

The architecture of the multi-modal interaction module is multi-layer transformers. The model sizes are listed in Table~\ref{table:model_size}.

\paragraph{Geographic Encoder}
All geographic feature embeddings are set to 256. The discretization factor $k$ is 10 and the grid number $N$ is 2000.
Geographic encoder has 4 layers of transformer with 256 hidden sizes. The mask probability is 0.15. The training batch size is 512. We use AdamW as optimizer with learning rate being 1e-4, weight decay being 0.02. We train geographic encoder for 30 epochs and take the last epoch checkpoint.

\paragraph{Pre-Training}
The training batch size is 512. We use AdamW as optimizer with learning rate being 5e-5, weight decay being 0.02. We train for 10 epochs and take the last epoch checkpoint.

\paragraph{Fine-Tuning}
For bi-encoder models, every training step has 56 queries, each has 20 candidates. We use AdamW as optimizer with learning rate being 5e-5, weight decay being 0.02. Specifically, ERNIE and StructBERT don't converge in this learning rate, we change it to 5e-6. We train geographic encoder for 10 epochs.

For cross-encoder models, every training step has 24 queries and the  learning rate for RoBERTa is 5e-6. Other settings are the same as bi-encoder.

\subsection{Ranking Results}

Table~\ref{table:ranking} gives the ranking results of both bi-encoder and cross-encoder PLMs. As the original StructBERT outperforms the other generic PLMs, it is used for further DA.
The generic PLMs directly fine-tuned on the downstream tasks show a low performance, which indicates that these two tasks are challenging.
Since cross-encoder can make fine-grained interactions among input features, while bi-encoder only interacts with the $CLS$ representations for the sake of efficiency, cross-encoder generally outperforms bi-encoder by a large margin.

By applying DA on bi-encoder, PLMs could gain an advantage over the generic ones. However, DA models consider only the textual modality and neglect the geographic modality. Through multi-modal pre-training, MGeo without query GC raises 2.37\% (\textit{resp.}, 1.07\%) point of Recall@1 on BERT-DA (\textit{resp.}, StructBERT-DA) by bridging the gap between query text and POI GC. After being accompanied by query GC, MGeo further shows a 3.55\% (\textit{resp.}, 1.77\%) improvement in Recall@1 over DA models with the help of incorporating correlations between query GC and POI text, as well as between query GC and POI GC. It is worth noting that GC of half the training and test queries use random locations to simulate the arbitrary geolocation of users, as described in Section~\ref{sec:geotes}. The results show MGeo is robustness and it may gain more improvements if the queries have more precise geolocations.

In cross-encoder, MGeo also shows superiority over baselines. DA brings fewer benefits on PLMs than it does in bi-encoder, \textit{i.e.}, 1.72\% on BERT and 0.14\% on StructBERT. However, improvements brought by incorporating the new geographic modal are consistent. MGeo without query GC gains 1.87\% (\textit{resp.}, 1.07\%) Recall@1 on BERT-DA (\textit{resp.}, StructBERT-DA). Together with query GC, MGeo boost DA models by 2.65\% (\textit{resp.}, 2.84\%) in Recall@1, showing effectiveness of multi-modal interaction.

\begin{table}[t]
    \caption{Ranking results of more baselines.}
    \label{table:more_baseline}
    \centering
    \small
    \begin{tabular}{llc}
        \toprule
                                       &                                                            & Recall@1       \\ \midrule
        \multirow{7}{*}{Bi-Encoder}    & DSSM~\cite{Zhao_Peng_Wu_Chen_Yu_Zheng_Ma_Chai_Ye_Qie_2019} & 34.59          \\
                                       & DPAM~\cite{Zhao_Peng_Wu_Chen_Yu_Zheng_Ma_Chai_Ye_Qie_2019} & 44.15          \\
                                       & PALM~\cite{Zhao_Peng_Wu_Chen_Yu_Zheng_Ma_Chai_Ye_Qie_2019} & 45.51          \\
                                       & BERT                                                       & 58.83          \\
                                       & ColBERT~\cite{10.1145/3397271.3401075}                     & 62.36          \\
                                       & Poly-Encoder~\cite{Humeau2020Poly-encoders:}               & 49.87          \\
                                       & BERT-MGeo                                                  & \textbf{76.04} \\ \midrule
        \multirow{3}{*}{Cross-Encoder} & BERT                                                       & 81.52          \\
                                       & ERNIE-GeoL~\cite{10.1145/3534678.3539021}                  & 82.94          \\
                                       & BERT-MGeo                                                  & \textbf{85.89} \\ \bottomrule
    \end{tabular}
\end{table}

\begin{table}[t]
    \caption{Retrieval results of bi-encoder.}
    \label{table:retrieval}
    \centering
    \small
    \begin{tabular}{lccc}
        \toprule
                  & BERT  & BERT-DA & BERT-MGeo      \\ \midrule
        Recall@1  & 21.70 & 51.76   & \textbf{52.70} \\
        Recall@5  & 29.32 & 60.82   & \textbf{63.39} \\
        Recall@20 & 35.70 & 67.08   & \textbf{70.49} \\
        Recall@50 & 40.30 & 71.61   & \textbf{75.00} \\
        MRR@5     & 24.58 & 55.29   & \textbf{56.79} \\ \bottomrule
    \end{tabular}
\end{table}

\subsubsection{More Baseline Comparisons}

Besides the PLM baselines, we also compare with more query-POI matching baselines, including two SOTA text-matching models, \textit{i.e.}, ColBERT~\cite{10.1145/3397271.3401075} and Poly-Encoder~\cite{Humeau2020Poly-encoders:}. ColBERT uses a late interaction architecture to enhance bi-encoder model. Similarly, Poly-Encoder uses attention mechanism to capture richer interactions between query and POI. Detailed introductions of DSSM, DPAM, and PALM can be found in~\cite{Zhao_Peng_Wu_Chen_Yu_Zheng_Ma_Chai_Ye_Qie_2019}. ERNIE-GeoL is a strong PLM cross-encoder baseline introduced in~\cite{10.1145/3534678.3539021}. Since the data and code of ERNIE-GeoL are not released, we only adopt the pre-training objectives. The results on the ranking task are shown in Table~\ref{table:more_baseline}. For bi-encoder, BERT-MGeo still outperforms ColBERT and Poly-Encoder, which capture more fine-grained interactions between query and POI. For cross-encoder, ERNIE-GeoL uses specific pre-training objectives to capture static geographic knowledge and outperforms BERT. While BERT-MGeo capture dynamic GC and outperforms ERNIE-GeoL.

\subsection{Retrieval Results}
Bi-encoder is also evaluated on the retrieval task, which focuses on finding the relevant POIs rather than ranking the correct POI at the top. Table~\ref{table:retrieval} reports Recall and MRR metrics.
Compared to BERT-DA, MGeo improves 3.41\% Recall@20. The results demonstrate that the effectiveness of MGeo in bi-encoder architecture stays consistent when the size of candidates becomes 100,000 times larger.

\subsection{Inference Time}

The inference time in seconds on 1 NVIDIA V100 GPU of bi-encoder and cross-encoder models is listed in Table~\ref{app:time}. For bi-encoder, we only count the time of query encoding, since the document can be encoded in advance in many industrial scenarios. We use 26 queries and 1040 documents for inference.

\begin{table}[t]
    \caption{Inference time in seconds.}
    \label{app:time}
    \centering
    \small
    \begin{tabular}{lcc}
        \toprule
                               & Bi-Encoder & Cross-Encoder \\ \midrule
        BERT-DA                & 0.0219     & 0.0396        \\
        BERT-MGeo w/o query GC & 0.0205     & 0.0414        \\
        BERT-MGeo              & 0.0269     & 0.0466        \\ \bottomrule
    \end{tabular}
\end{table}

\subsection{Ablation Study}

Since we use the same bi-encoder models for both retrieval and ranking tasks, the ablation study is mainly conducted on ranking task of BERT-based models.

\begin{table}[t]
    \caption{Influence of different geographic object types.}
    \label{table:geo_ele_tp}
    \centering
    \small
    \begin{tabular}{lcccc}
        \toprule
                & \multicolumn{2}{c}{Bi-Encoder} & \multicolumn{2}{c}{Cross-Encoder}                                   \\ \midrule
                & Recall@1                       & MRR@5                             & Recall@1       & MRR@5          \\ \midrule
        Line    & \textbf{74.56}                 & \textbf{82.57}                    & 83.71          & 88.88          \\
        Polygon & 74.26                          & 82.51                             & \textbf{84.84} & \textbf{89.85} \\ \midrule
        Both    & \textbf{76.04}                 & \textbf{83.85}                    & \textbf{85.89} & \textbf{90.74} \\ \bottomrule
    \end{tabular}
\end{table}

\subsubsection{Geographic Object}
\begin{figure}[t]
    \centering
    \includegraphics[width=.6\linewidth]{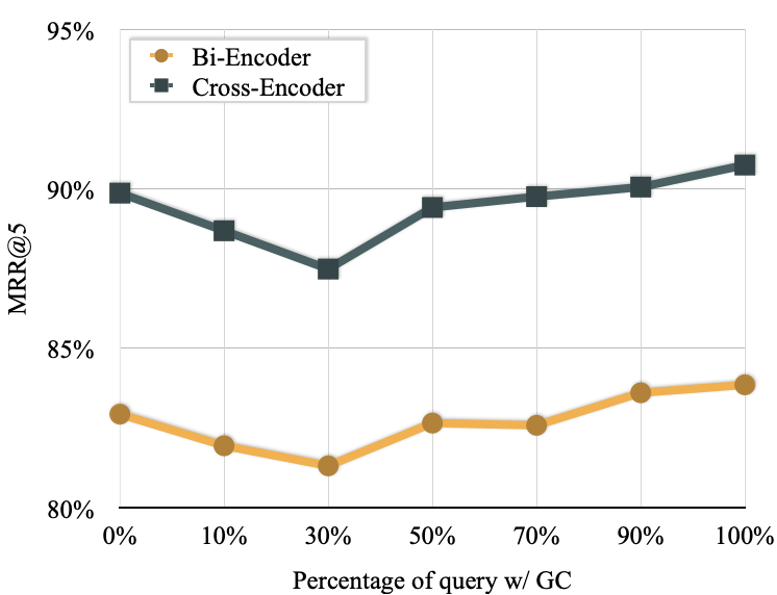}
    \caption{Ranking MRR@5 for different percentage of queries that have GC.}
    \label{fig:qgi}
\end{figure}

We first study the influence of training queries with GC. We randomly remove GC of the same proportion from the training, development, and test queries. As shown in Figure~\ref{fig:qgi}, the performance is impaired when a small proportion of queries contain GC. This decrease comes from a larger proportion of noise. Taking 30\% of queries having GC as example, there are already 15\% of GC inputs are random~(half GC are randomly selected). Since it is difficult to distinguish query without GC from query without geographic object~(but with geolocation), the rest queries without GC can be considered as noise too. Thus we have in total 75\% queries with noisy GC, which damages model performance. When noises proportion becomes smaller than 65\%~(70\% query with GC), the performance is better than training without query GC.

The influence of different geographic object types is reported in Table~\ref{table:geo_ele_tp}. There is not a huge gap between line and polygon for bi-encoder, while cross-encoder can perform better with only polygon than only line, as there are more polygons than lines in the GIS. This also suggests that cross-encoder is better at capturing the fine-grained correlations than bi-encoder. Nevertheless, using either line or polygon is better than the single-modal baselines. Besides, bi-encoder and cross-encoder can have a better performance when the two types of geographic objects both present.

\subsubsection{Query Type}
\begin{figure}
    \centering
    \begin{subfigure}[t]{.49\linewidth}
        \centering
        \includegraphics[width=1.\linewidth]{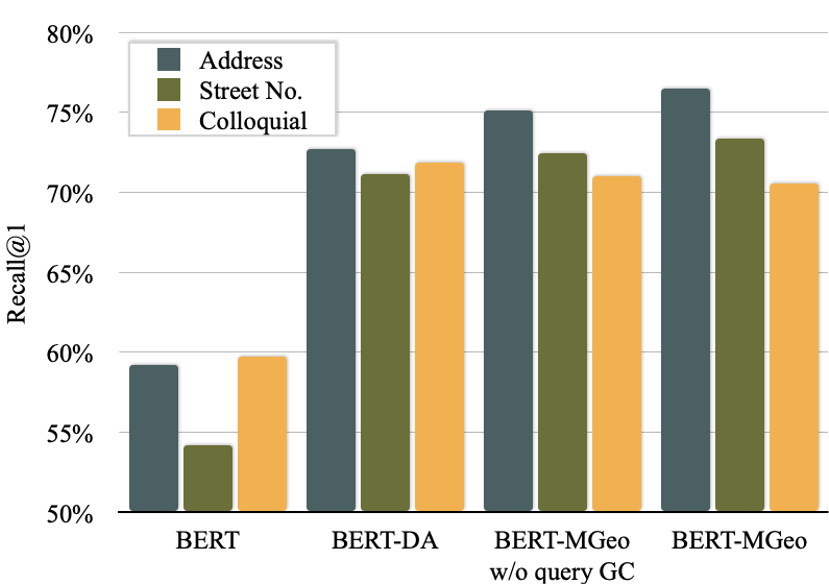}
        \caption{Bi-Encoder.}
        \label{fig:qtp_bi}
    \end{subfigure}
    \begin{subfigure}[t]{.49\linewidth}
        \centering
        \includegraphics[width=1.\linewidth]{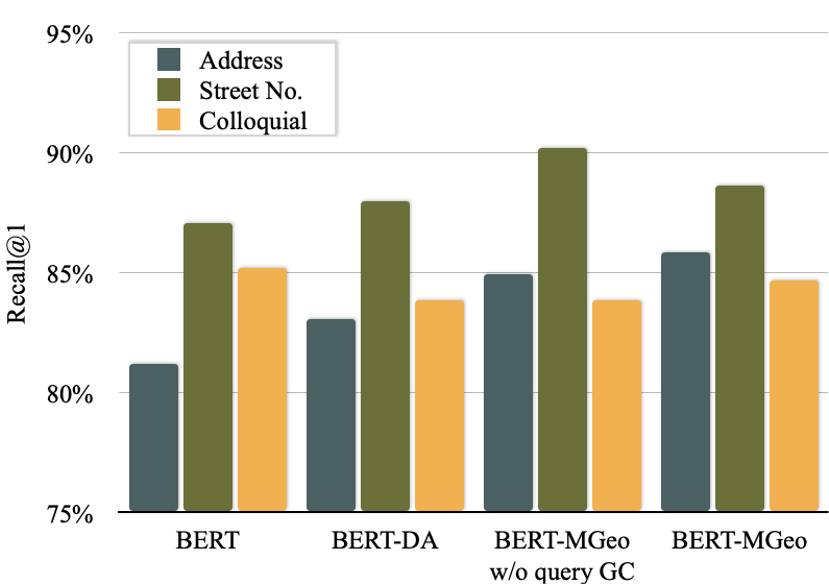}
        \caption{Cross-Encoder.}
        \label{fig:qtp_cross}
    \end{subfigure}
    \caption{Ranking Recall@1 for (A) bi-encoder and (B) cross-encoder with different query types.}
    \label{fig:qtp}
\end{figure}

Figure~\ref{fig:qtp} shows the performance on three query types, \textit{i.e.}, address, street number, and colloquial.
Bi-encoder models perform best on address description, while cross-encoder models perform best on street number description. This suggests that cross-encoder is better at capturing fine-grained correlations. Colloquial query contains many daily expressions, which rarely appear in domain corpus. Thus BERT-DA is even worse than BERT on it. However, the use of GC help reduce this shortcoming of DA.

\subsubsection{Amount of Training Data}
\begin{figure}
    \centering
    \begin{subfigure}[t]{.49\linewidth}
        \centering
        \includegraphics[width=1.\linewidth]{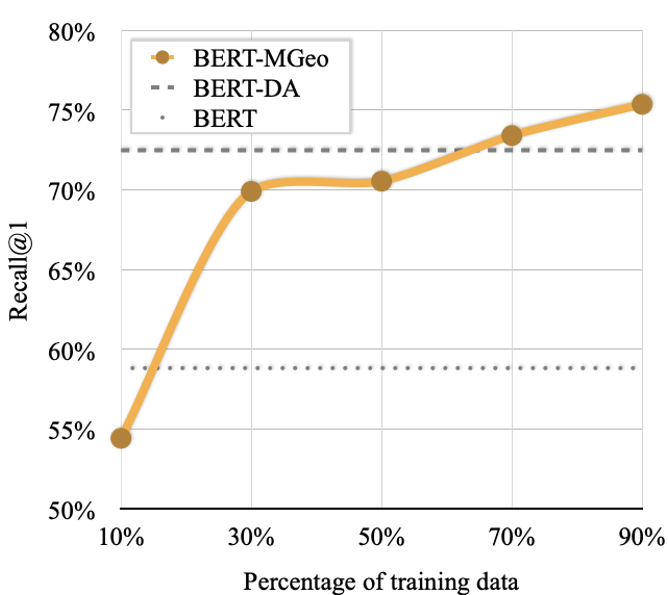}
        \caption{Bi-Encoder.}
        \label{fig:pct_bi}
    \end{subfigure}
    \begin{subfigure}[t]{.49\linewidth}
        \centering
        \includegraphics[width=1.\linewidth]{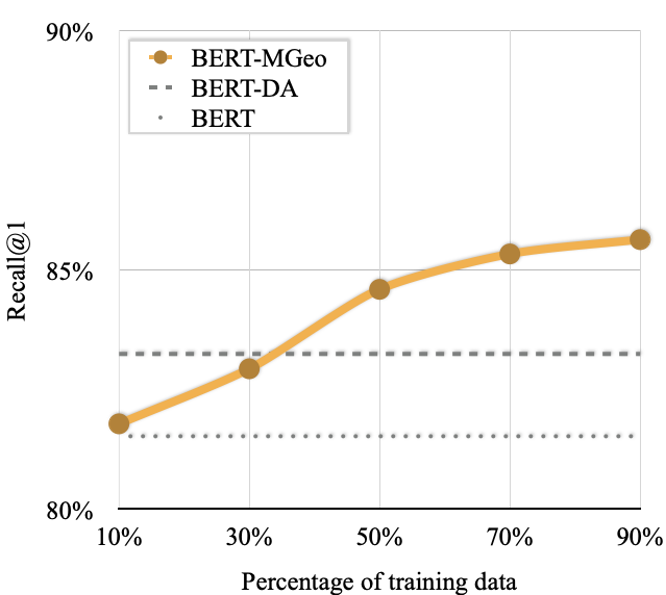}
        \caption{Cross-Encoder.}
        \label{fig:pct_cross}
    \end{subfigure}
    \caption{Ranking Recall@1 for (A) bi-encoder and (B) cross-encoder with different amounts of training data.}
    \label{fig:pct}
\end{figure}

We study the performance of MGeo with different amounts of training data. As shown in Figure~\ref{fig:pct}, the dashed line is used for representing BERT-DA and the dotted line for the original BERT. With only 30\% of training data, the bi-encoder and cross-encoder using MGeo can outperform the BERT baseline by a large margin.

\subsubsection{Query Incompleteness}

\begin{figure}
    \centering
    \begin{subfigure}[t]{.49\linewidth}
        \centering
        \includegraphics[width=1.\linewidth]{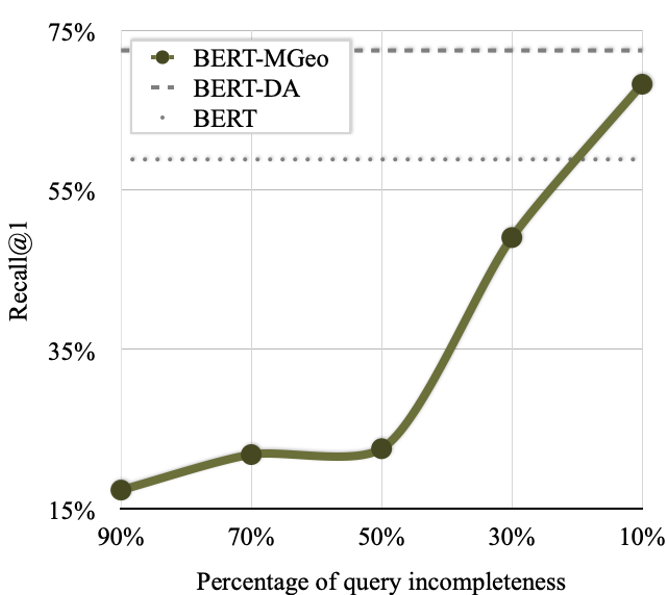}
        \caption{Bi-Encoder.}
        \label{fig:qic_bi}
    \end{subfigure}
    \begin{subfigure}[t]{.49\linewidth}
        \centering
        \includegraphics[width=1.\linewidth]{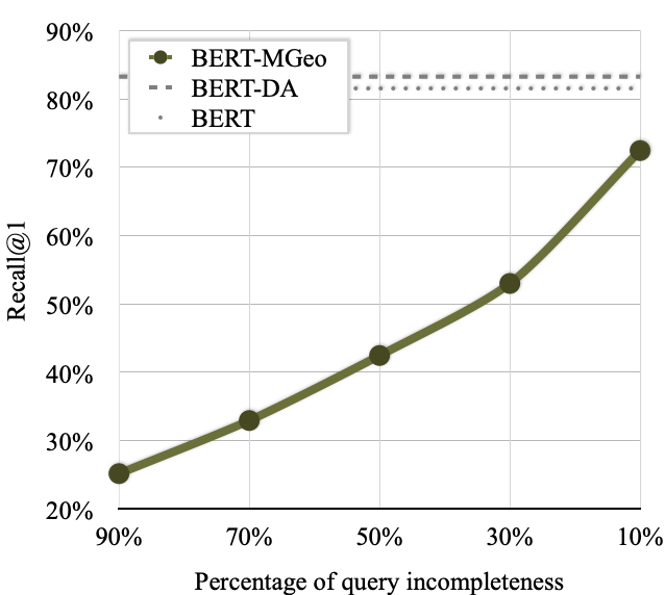}
        \caption{Cross-Encoder.}
        \label{fig:qic_cross}
    \end{subfigure}
    \caption{Ranking Recall@1 for (A) bi-encoder and (B) cross-encoder with different percentage of query incompleteness.}
    \label{fig:qic}
\end{figure}

POI suggestion also plays an important role in LBS, where the name of POIs are listed when the input is unfinished. To simulate such scenario, we also evaluate MGeo on incomplete queries by truncating the trailing characters.
Figure~\ref{fig:qic} shows the performance with different truncation ratio of the test queries. The results demonstrate that bi-encoder using MGeo could outperform the BERT baseline with full queries with a small truncation ratio. Whereas the cross-encoder could not, since the semantic similarity is more important for cross-encoder.

\section{Conclusion}

In this paper, we formalize the important concept of \textit{Geographic Context}~(GC), which is indispensable for real-world human POI exploration process. We propose a multi-modal geographic language model MGeo, which considers GC as a new modality. Therefore, GC can be represented together with text. In addition, we build a new open-source large-scale benchmark GeoTES to facilitate further research on the query-POI matching topic. Extensive experiments are conducted to evaluate our proposed method on the state-of-the-art PLMs, and the detailed analyses demonstrate that MGeo can significantly outperform other baselines. Even though geolocation of user may be absent and query has no GC, MGeo can still obtain improvements over the baselines, showing its capability of modeling text-to-text, GC-to-GC and text-to-GC correlations. For future work, other modalities like POI image can be further explored, as well as more inventive geographic encoder. Besides, our proposed GC modeling has the potential to boost all geography-related tasks.

\bibliographystyle{ACM-Reference-Format}
\bibliography{base_series}

\end{document}